\documentclass[letterpaper, 10 pt, conference]{ieeeconf}  

\IEEEoverridecommandlockouts                              
\usepackage{amsmath}
\usepackage{amsfonts,amssymb}
\usepackage{bm}
\usepackage{graphicx}
\usepackage{subcaption}
\usepackage[linesnumbered,ruled]{algorithm2e}
\usepackage{balance}
\usepackage{array}
\usepackage{color}
\usepackage{hyperref}
\usepackage{multirow}

\usepackage{bm}
\newcommand{\vect}[1]{\boldsymbol{\mathbf{#1}}}

\newcommand{\etal}{\textit{et al.}}

\title{\LARGE \bf Sparse2Dense: From direct sparse odometry to dense 3D reconstruction}
\author{Jiexiong Tang$^{1}$, John Folkesson$^{1}$ and Patric Jensfelt$^{1}$
\thanks{This work was partially supported by the Wallenberg AI, Autonomous Systems and Software Program (WASP) and the SSF project FACT.} 
\thanks{$^{1}$The authors are all with the Centre for Autonomous Systems at KTH Royal Institute of Technology, Stockholm, SE-10044, Sweden 
{\tt\small jiexiong@kth.se}}%
}

\begin{document}

\thispagestyle{empty}
\onecolumn
\textcopyright 2019 IEEE.  Personal use of this material is permitted.  Permission from IEEE must be obtained for all other uses, in any current or future media, including reprinting/republishing this material for advertising or promotional purposes, creating new collective works, for resale or redistribution to servers or lists, or reuse of any copyrighted component of this work in other works.

\twocolumn
\newpage

	\maketitle
	\thispagestyle{empty}
	\pagestyle{empty}

\begin{abstract}
	
	In this paper, we proposed a new deep learning based dense monocular SLAM method. Compared to existing methods, the proposed framework constructs a dense 3D model via a sparse to dense mapping using learned surface normals. With single view learned depth estimation as prior for monocular visual odometry, we obtain both accurate positioning and high quality depth reconstruction. 
	The depth and normal are predicted by a single network trained in a tightly coupled manner.
	Experimental results show that our method significantly improves the performance of visual tracking and depth prediction in comparison to the state-of-the-art in deep monocular dense SLAM.
\end{abstract}

\section{Introduction}
{SLAM} is a key building block in most mobile autonomous systems. Much of the recent research addresses the SLAM problem with a single camera. A solution with a single camera would be very competitive in many applications as a camera is relatively inexpensive and already present in most mobile devices. In this paper we investigate direct methods for SLAM. Impressive semi-dense/sparse tracking and mapping results have been achieved. LSD-SLAM~\cite{LSD-SLAM} and the more recent DSO~\cite{DSO} define the state of the art in these domains. However, they are not able to overcome the intrinsic problem of monocular visual positioning, that scale is not observable. With the recent advances in deep learning, this issue is now being tackled by using learning based depth estimators. The idea is to use a network to predict the depth from a monocular image and use this as a prior in a SLAM or visual odometry (VO) system. Recent works~\cite{CNN-SLAM,DVSO} show that the absolute position error can be greatly reduced in this way. This is the approach we take in this paper as well.

\begin{figure}[!htp]
	\centering
	\begin{subfigure}{0.45\textwidth}
		\centering
		\includegraphics[width=0.95\textwidth]{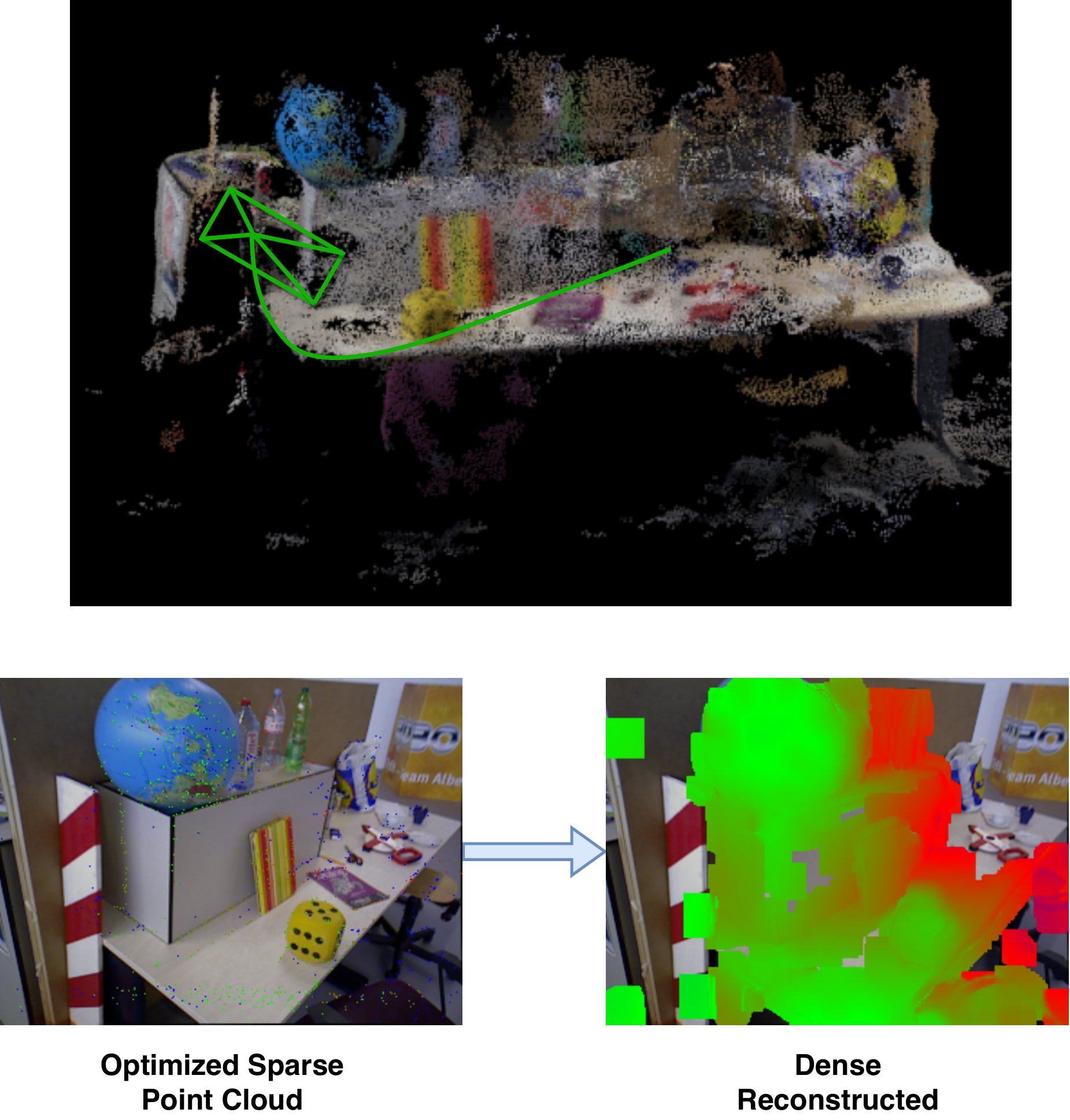}
	\end{subfigure}
	\caption{An example of our proposed method, S2D, for dense reconstruction. Top: The reconstructed 3D scene.  Left: The optimized sparse point cloud overlaid on the corresponding image frame. Right: The densely reconstructed point cloud.
	Note how the depth of the points in the local region around the sparse points are refined.}
	\label{fig:digest}
    \vspace*{-0.25cm}
\end{figure}


Our long term goal is accurate and dense 3D reconstruction of scenes. Such models could, for example, support advanced predictions of the effect of certain physical interactions. We make several important contributions in this paper. At a high level, we propose a deep learning based dense monocular SLAM method capable of real-time performance. The most related work to ours is CNN-SLAM~\cite{CNN-SLAM}. A key insight in our work is that we should combine the ability of the CNN to generate dense depth predictions with the ability of a visual tracking system to generate highly accurate but sparse points through optimization. We use these sparse but accurate points to correct the dense depth predictions from the CNN. In particular, we leverage normals and an assumption about local planar structures. Depth and normals are predicted by a single CNN for efficiency. The network has been trained in a novel coupled manner, optimized for the sparse-to-dense reconstruction task. The CNN thus contributes to the tracking system by providing the true scale and the tracking system helps improve the accuracy of the dense depths. In our work, the sparse point clouds are generated from the active window keyframes of DSO~\cite{DSO}, initialised by depth priors from the CNN. After the sparse to dense reprojection, the keyframes are sent to the backend which includes keyframe wise refinement and a fusion based mapper. 
We choose a fusion based mapper for the further enhancement of the global 3D model consistency. The backend also provides our system with loop closure abilities. An example is shown in Fig.~\ref{fig:digest}.

In summary, the key feature of our system is learning based sparse to dense mapping for 3D reconstruction. We denote our method S2D (Sparse2Dense). In the remainder of the paper, we firstly review the related work. Secondly, we provide an overall system overview to better explain the details. Then, in the following two sections, we present our approach to coupled training and reconstruction. Finally, we show experimental results, discuss these and outline directions for future work.

\section{Related Works}

\textbf{Single View Depth and Normal Estimation.}
Deep learning methods have achieved great advances in the area of single view depth and/or normal estimation and have largely replaced classical methods such as~\cite{Make3d} and~\cite{pop-up}.

Eigen~\etal~\cite{eigen1} train a two scale CNN to predict depth from single images.
Liu~\etal~\cite{neural-field15} use a CNN to learn unary and pairwise potentials for a continuous CRF for depth estimation. 
Laina~\etal~\cite{FCRN} propose a fully convolutional~\cite{FCN} residual network~\cite{ResNet}~(FCRN) with up-projection based upsampling using interleaved convolution. 
In addition, there are many supervised deep learning methods for monocular depth estimation \cite{depth-1, depth-2, depth-3, depth-4} showing good performances. 

Another recent trend, are approaches which train the CNN to predict the depth in a self-supervised way \cite{Garg,Deep3D,Left-right,Single-stereo} or in an unsupervised way \cite{SFMLearner}.  They use an image reconstruction loss without the supervision of ground truth depth. This is well suited for scenarios where the depth ground truth is hard or expensive to be collected, e.g., recorded video and outdoor. 
Recent methods in~\cite{semo,DVSO} show that combining  supervised learning using depth ground truth and self-supervised learning achieves better performance. 

For single view normal prediction, Wang~\etal~\cite{Wang} developed CNNs that operate both locally and globally on the image. The resulting predictions are combined with evidence from vanishing points to produce the final prediction.
In~\cite{eigen2} both depth and normal predictions are performed using a multi-scale deep network.
Li~\etal~\cite{depth-crf} use hierarchical CRFs to estimate depth and normals from monocular images. 
Bansal~\etal introduce a skip-network model in~\cite{2d-3d} and in~\cite{PixelNet} a model for stratified sampling of pixels that can be used for normal prediction.
In work~\cite{Surge}, a CRF with a 4-stream CNN is designed to improve the consistency of predicted depth and surface normals in planar regions.
GeoNet, proposed in recent work ~\cite{GeoNet}, consists of two streams of CNNs that have been jointly optimized to predict depth and normal through depth-to-normal and normal-to-depth mappings. 

In our setting, an indoor scenario, relative large amounts of labeled samples is available, such as NYUv2~\cite{NYUv2}, SUN-3D~\cite{SUN-3D}, etc. Thus, we trained the network in a supervised manner. Furthermore, we trained the CNN to predict both depth and normal in a coupled way similar to~\cite{GeoNet}. The reconstructed depth and normals are used as strong regularization and can be seen as a pre-optimization for the sparse to dense reconstruction. Further details can be found in Sec.~\ref{sec:coupled_depth_and_normal_prediction}.

\textbf{Monocular VO and SLAM} 
Impressive progress has been made in visual odometry and SLAM methods. A common way to categorize different approaches is to use direct / indirect and dense / sparse. In direct methods, image frames are aligning directly based on pixel intensities and in indirect methods by first extracting features. Sparse and dense methods differ by how much of the image information is used. ORB-SLAM~\cite{ORB-SLAM2} defines the state-of-the-art in indirect sparse methods. When speed is of the essence SVO2~\cite{SVO}, using a  semi-direct approach, offers frame rates of hundreds of Hz. LSD-SLAM~\cite{LSD-SLAM} was one of the first direct semi-dense methods. The more recent DSO~\cite{DSO} is a direct and sparse method that adds joint optimization of all model parameters. 

Scale drift \cite{scale-drift} is an error which cannot be removed easily in a principled way with traditional methods when using a single camera\footnote{Observing objects with known sizes has been one way to overcome scale-drift.}. Traditional, non-deep, methods are therefore gradually being challenged by learning based methods.
Recent deep learning based mapping systems \cite{CNN-SLAM, DVSO} reduce the scale drift error by incorporating deep learning based single view depth estimation.
In CNN-SLAM \cite{CNN-SLAM}, a CNN is used to predict single view depth, which is fed into LSD-SLAM to achieve dense reconstruction. The depth is refined by using Bayesian filtering from \cite{LSD-SLAM, LSD-VO}.
In DVSO \cite{DeepVO}, a virtual stereo view similar to \cite{Left-right} is predicted for the depth. This is jointly optimized for high accuracy tracking using DSO.
Yin~\etal~\cite{scale-recovery} improve the performance of the depth
estimation by using two consecutive frames and estimate ego-motion with refined depth.
In CodeSLAM \cite{code-slam}, a compact learned representation from conditioned auto-encoding is optimized to obtain a dense reconstruction with camera pose.

End-to-end training is a general trend. Here ego-motion estimation is performed directly, either supervised with ground truth or unsupervised~\cite{SFMLearner, Undeepvo} using image reconstruction loss. However, as shown in \cite{DeepVO}, the performance of the end-to-end ego-motion is not on par with geometrical optimization based methods yet.

Our work is tightly related to deep learning based VO/SLAM and single view depth/normal estimation. Our method, S2D, is built on top of the direct monocular VO method DSO\cite{DSO}. Depth and normals are predicted by a jointly optimized CNN.
The learning based depth prior is used in the geometric optimization to reduce scale drift and achieve accurate monocular camera pose estimation.
This results in sparse but optimized depth estimates.
Finally, surface normal based geometrical reconstruction is conducted to rebuild a dense point cloud from the optimized sparse depth estimates.

\section{System Overview}

\begin{figure*}[!ht]
	\centering
	\begin{subfigure}{0.95\textwidth}
	\includegraphics[width=0.95\textwidth]{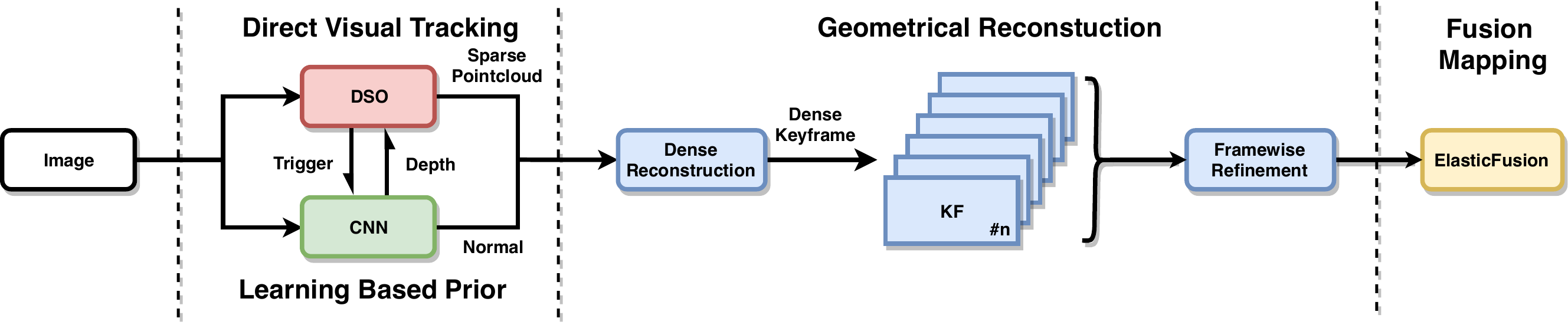}
	\end{subfigure}
	\caption{An overview of our sparse to dense mapping framework, S2D.}
	\label{fig:pipe_line}
	\vspace*{-0.5cm}
\end{figure*}

\begin{figure*}[!htp]
    \centering
    \begin{subfigure}{0.24\textwidth}
    \includegraphics[width=\textwidth]{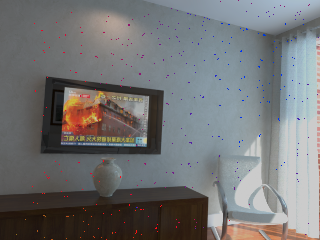}
    \caption{}
    \end{subfigure}
    \centering
    \begin{subfigure}{0.24\textwidth}
    \includegraphics[width=\textwidth]{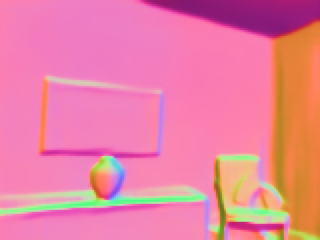}
    \caption{}
    \end{subfigure}
    \centering
    \begin{subfigure}{0.24\textwidth}
    \includegraphics[width=\textwidth]{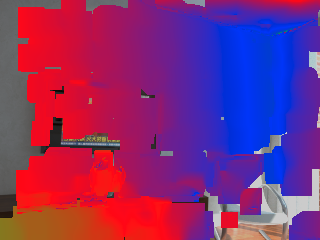}
    \caption{}
    \end{subfigure}
    \centering
    \begin{subfigure}{0.24\textwidth}
    \centering
    \includegraphics[width=\textwidth]{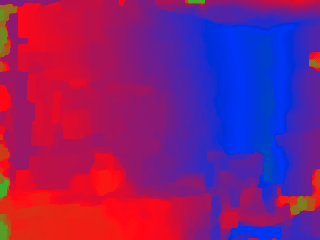}
    \caption{}
    \end{subfigure}
	\caption{The sparse to dense procedure, figures above progressively shows the intermediate outputs: (a) optimized sparse depth image using CNN depth as prior; (b) CNN normal; (c) dense reconstruction using (a) and (b); (d) after (c) has been refined with adjacent keyframes.}
	\label{fig:sparse_to_dense}
	\vspace*{-0.25cm}
\end{figure*}

The overview of proposed S2D system is shown in Fig.~\ref{fig:pipe_line}. The overall framework can be divided into four major stages: learning based prior generation for depth/normal, visual tracking using direct alignment, geometrical sparse to dense reconstruction and lastly fusion based mapping. The main contributions in this paper are made in stage one and three. Examples of intermediate results in our pipeline are illustrated in Fig.~\ref{fig:sparse_to_dense}.

Before we dive into the fine technical details, we provide a brief overview of these four stages and how they are connected. We use DSO for the visual tracking. Whenever a new keyframe is created by DSO, we use a single network to infer the depth/normals from the image data. When DSO is in the initialization stage, we directly assign depth priors to the new immature points to be optimized. If, on the other hand, DSO has been initialized, we project all mature points from active keyframes in the optimization window to the new keyframe. This sparse optimized depth cloud is used for: (1) a global scale correction for the depth prior of the new immature points and (2) a sparse to dense recasting, where the optimized depth is propagated to co-planar neighbouring pixels using the predicted normal. The reconstructed depth images of theses keyframes are refined using Bayesian estimation. Finally, the dense refined depth images are fed into a fusion based mapper, built on ElasticFusion, which generates a consistent global 3D model and handles loop closures.

\section{Coupled Depth and Normal Prediction}
\label{sec:coupled_depth_and_normal_prediction}

\begin{figure}[!t]
	\centering
	\includegraphics[width=0.8\linewidth]{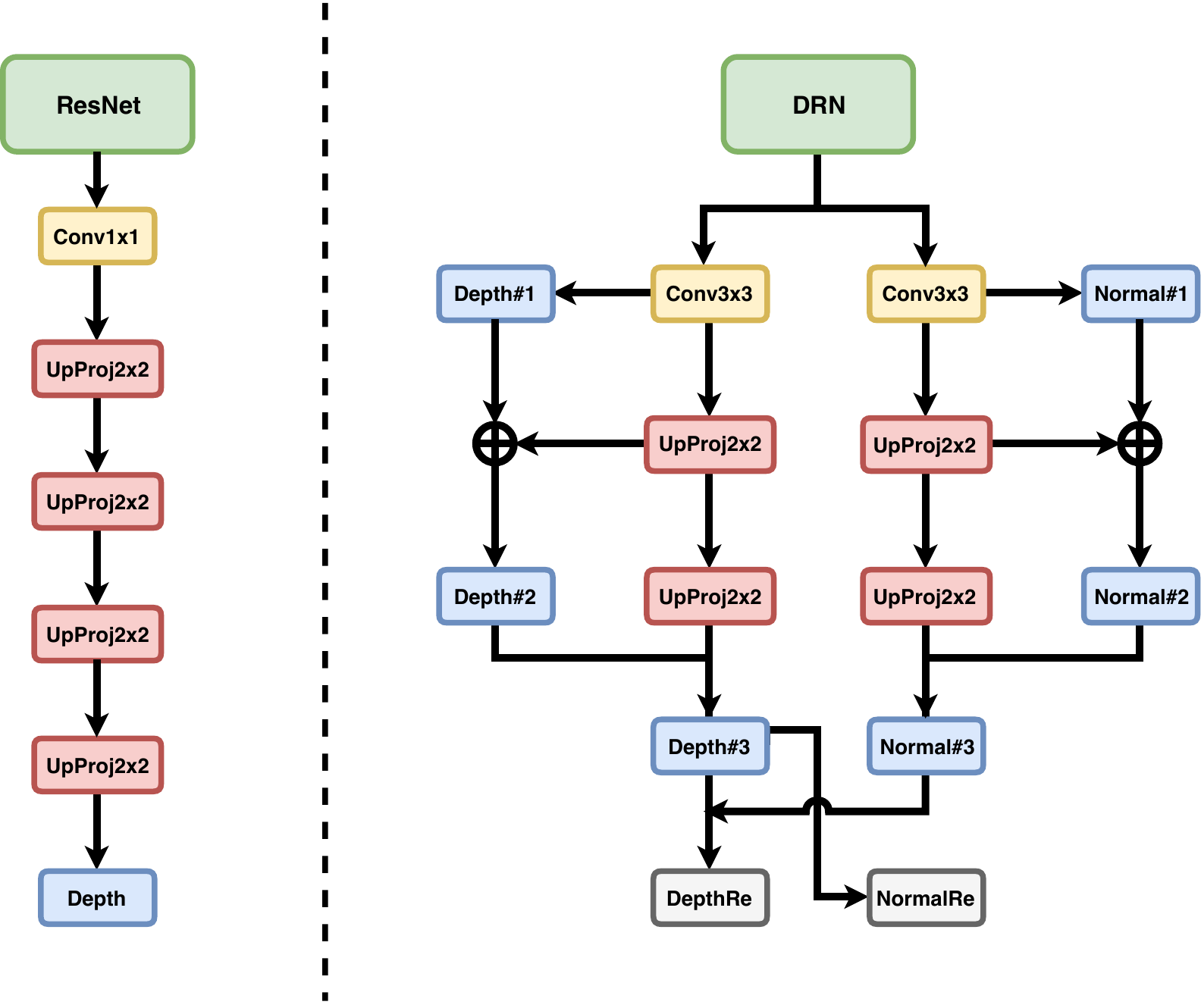}
	\caption{Network structure comparison of the original FCRN (left) and our FCDRN (right).}
	\label{fig:network_structure}
\end{figure}
In this section, we first introduce the network structure deployed in S2D. Then, we illustrate the training scheme for the tightly coupled depth and normal prediction.

\subsection{Network Structure}
\label{sec:FCDRN}
In CNN-SLAM, the network FCRN with an encoder-decoder structure is used for monocular depth estimation. The encoder part is built on ResNet50. For the decoder, a sequence of residual upsampling blocks, composed of interleaved convolution and up-projection, are used for recovering the feature maps at the targeted resolution. 
In S2D, various modifications have been made for more stable training and better performance. A source of inspiration is \cite{Left-right}, which present a structure that obtain better performance than related methods (including FCRN) in outdoor scenarios. With respect to FCRN, we replace ResNet50 by the Dialated Residual Network~(DRN)~\cite{DRN} in the encoder part. The feature maps of the DRN have higher resolution, which is better suited for generating more detailed depth/normal. For the decoder, besides residual upsampling, we train the network to predict depth/normal at three different scales. The depth prediction at lower resolution are upsampled and aggregated with the higher resolution one during the decoding. The overall structure is shown in Fig.~\ref{fig:network_structure}. We denote the modified network structure FCDRN, to highlight the interleaved upsampling block from FCRN and the use of DRN.

\subsection{Training Scheme}

Inspired by GeoNet~\cite{GeoNet}, we use a tightly coupled two-way reconstruction training scheme: depth-to-normal and normal-to-depth. \cite{GeoNet} showed impressive depth and normal estimation quality compared with other existing methods. GeoNet uses two CNNs to predict the depth/normal. It runs at around 1Hz on a desktop with modern CPU and GPU, which is much more efficient than other related deep learning based methods (more than 10 times faster). However, an online SLAM system can still not afford this computational cost.

In S2D, the depth/normal are predicted in one single network (FCDRN introduced in Sec.~\ref{sec:FCDRN}) using input images with resolution of $320\times240$. This is similar to the resolution of $304\times228$ used in FCRN. The output, the depth/normal predictions, is $160\times120$. 
We do not use depth-to-normal mapping as post-processing, as we found its main effect to be to regularize the depth with geometrical structure during training, but not to improve the quality of the predicted depth.

We implement the normal-to-depth conversion in C++ with CUDA to allow it to run in real time together with the whole system. 
Note that the overhead of computing depth/normal only appears when making a new keyframe.

\textbf{Focal Length Adaption.} The main challenge when training a network to predict depth from RGB-D images captured by a single RGB-D camera is: if the testing is conducted using another sensor, the change in focal length brings in an error in the scale of the estimated depth. To reduce this effect and make the trained network generalize better, in CNN-SLAM \cite{CNN-SLAM}, the depth $\hat{\vect{Z}}$ generated by the CNN is adjusted as follows:
\begin{equation} \label{eq:depth_adapt1}
\vect{Z}(\vect{u}_i) = \frac{f_{test}}{f_{train}} \hat{\vect{Z}}(\vect{u}_i)
\end{equation}
where ${f_{train}}$ and ${f_{test}}$ are the focal lengths of the cameras used for training and testing respectively. 
This rescaling is performed as a separate post-processing step, which is not related to the training of the network. In S2D, we choose the disparity as the target for the CNN to regress rather than the depth. By doing so, we embed the scale correction into the training to better diminish the effect mentioned above. The depth is calculated from the disparity $\hat{\vect{D}}$ as follows:
\begin{equation} \label{eq:depth_adapt2}
\vect{Z}(\vect{u}_i) = \frac{B \; f_{train}}{\hat{\vect{D}}(\vect{u}_i)}
\end{equation}
where $B$ is a hyperparameter that can be seen as a ``virtual'' baseline. It controls the range of the depth to be regressed and is set to $0.1$m in our implementation. Note that the disparity is linearly dependent on the inverse depth, which is well-known to have various statistical advantages and also converge better in our optimization. 
By using Eq.\ref{eq:depth_adapt2}, we decouple the focal length from the training. More importantly, the network now predicts the disparity from a fixed base line camera rather than a camera with a fixed focal length. 

\textbf{Depth-to-Normal}. 
The depth-to-normal conversion is straightforward. As mentioned, the original least mean square estimation is slow. To accelerate the training, we adopt a common approach from SLAM and compute the normal using the cross product~\cite{ElasticFusion}: 
\begin{equation} \label{eq:normal_from_depth}
\begin{gathered}
\resizebox{1.0\hsize}{!}{$\vect{N}_{re}(\vect u_i) = \nu [ \vect V( u_i + 1,  v_i) - \vect V(u_i, v_i) ) \times (\vect  V(u_i, v_i + 1) - \vect V(u_i, v_i) ) ] $ } \\
\resizebox{0.8\hsize}{!}{$
	\vect V(u_i, v_i) = [x_i, y_i, z_i]^T = [ \frac{z_i(u_i - c_x )}{f_x}, \frac{z_i(v_i - c_y )}{f_y}, z_i]^T$ }
\end{gathered}
\end{equation}
where $\nu[\vect n] = \vect n / ||\vect n||_2$ and $\vect V(u,v)$ is the 3D vertex un-projected from image plane. 
Compared to the least mean square estimation used in \cite{GeoNet}, we did not find a notable difference in the quality of depth/normal estimation. 

\textbf{Normal-to-Depth.} The normal-to-depth conversion is based on the assumption that points used to reproject the depth to the current position are locally on the same surface tangent:
\begin{equation} \label{eq:depth_from_normal_1}
n_{xi} (x - x_i) +  n_{yi} (y - y_i) + n_{zi} (z - z_i) = 0
\end{equation}
Rearranging the equation and substituting $x$ with a coplanar point $x_j$: 
\begin{equation} \label{eq:depth_from_normal_2}
\begin{gathered}z_{ij} = \frac{n_{xj} x_j + n_{yj} y_j + n_{zj} z_j  }{ \frac{(u_i - c_x) n_{xj}}{f_x} + \frac{(v_i - c_y) n_{yj}}{f_y} + n_{zj}}
\end{gathered}
\end{equation}
The above equation shows how the depth can be reprojected by using the normal and depth of neighbouring points. The final depth can then be computed using the weighted sum of every depth reprojected by the neighbouring points. The weighted sum for each pixel corresponds to a spatial filter for which the kernel weights are given by the inner products of normals of neighbouring points:
\begin{equation} \label{eq:depth_from_normal_3}
\vect Z_{re}(\vect u_i) = \frac{\sum_{j\in \vect{\mathcal{C}}_i} \vect n_j^T \vect n_i z_{ij}}{\sum_{j\in \vect{\mathcal{C}}_i} \vect n_j^T \vect n_i }
\end{equation}
where $\vect{\mathcal{C}}_i$ contains the pixels around $u_i$ meeting the following condition:
\begin{equation} \label{eq:depth_from_normal_4}
\vect{\mathcal{C}}_i = \{(x_j , y_j , z_j )| \vect n_j^T \vect n_i > \psi, |u_i - u_j | < \sigma, |v_i - v_j | < \sigma\}
\end{equation}
where $\psi$ is a threshold to remove non-coplanar points and $\sigma$ is the spatial distance in the image plane. To be consistent with \cite{GeoNet}, they are set as $0.95$ and $5$ respectively.

\textbf{Objective Functions.} The overall objective function is as follows:
\begin{equation} \label{eq:loss_all}
\begin{gathered}
\mathcal{L}(\vect u_i) = \underbrace{ \alpha \; ( ||\hat{\vect D}(\vect u_i) -\vect D^{gt}(\vect u_i)||_{\epsilon} + ||\hat{\vect N}(\vect u_i) - \vect N^{gt}(\vect u_i)||_{\epsilon} )}_{\text{Supervised Regression}} \\
\underbrace{ +  \beta \; ||\vect D_{re}(\vect u_i) - \vect D^{gt}(\vect u_i)||_{\epsilon} + \gamma \; ||\vect N_{re}(\vect u_i) - \vect N(\vect u_i)||_{1}}_{\text{Coupled Refinement}}
\end{gathered}
\end{equation}
where $\vect D_{re}$ is computed from $\vect Z_{re}$ via Eq.~\ref{eq:depth_adapt2} to be in the same scale as $\hat{\vect D}$ and ${\vect D}^{gt}/\vect{N}^{gt}$ are the ground truth disparity and normals respectively, $||.||_{\epsilon}$ is the huber loss and $||.||_{1}$ is the L1 loss. The threshold for the huber is set as $0.1$ relative to the maximum absolute error. The hyper-parameters $\{ \alpha, \beta, \gamma \}$ control the weights of the different terms and are set to $\{1.0, 0.1, 0.05 \}$ in our implementation. The supervised term is the ordinary regression using ground truth disparity and normal. The coupled terms consists of the penalty on the reconstructed normal and depth. The supervised term is given higher weight because the corresponding information is more reliable. Henceforth, we denote the depth and normal predicted by the CNN as CNN depth and CNN normal to avoid confusion.

\section{SLAM}

In this section, we introduce the tracking frontend, deep sparse visual odometry, and the mapping backend, dense global fusion.

\subsection{Deep Sparse Visual Odometry}

For the visual odometry, our implementation is based on DSO~\cite{DSO}\footnote{\url{https://github.com/JakobEngel/dso}}.	
In S2D, the CNN depth is used as a prior for the initialization of sparse points. 
We do not force DSO to initialize densely for the following reason: even if points in flat regions are activated using CNN depth, the uncertainty of those are unlikely to be reduced during the tracking since the gradients they contribute are relative small. In fact, we found that the tracking performance degrades if large amounts of low-gradient points are forced into the joint optimization.

\textbf{Online Scale Correction.} When a new key frame is required by DSO, we warp all visible mature points from active keyframes into the current image plane:
\begin{equation} 
\label{eq:depth_optimized_1}
\begin{aligned}
\vect Z_{opt}({\vect u_i}) &=  z_i^* \\
[u_i^*, v_i^*, z_i^*]^T &=  V^{-1}( \vect R \; \vect V(u_i, v_i) + \vect t)
\end{aligned}
\end{equation}
where $[\vect{R}, \vect{t}] \in \mathbb{SE}(3)$ are the relative transforms between active keyframes and the new keyframe. They are estimated by direct image alignment of DSO. $\vect Z_{opt}$ is the warped optimized sparse depth image, it has been corrected in range and structure by tracking.
We perform online scale corrections to the CNN depths using the scale changes observed in the sparsely optimized point:
\begin{equation} 
\label{eq:depth_optimized_2}
{\vect Z_{cor}}({\vect u_i}) = \vect Z(\vect u_i) \frac{\sum_{j \in \vect{\Omega} } B^{rel}_j \; {z_j^*} / {z_j}     }{\sum_{j \in \vect{\Omega}}B^{rel}_j} \
\end{equation}
where $B^{rel}$ is the maximum relative baseline from which the point has been observed. 
$\vect{\Omega}$ includes all mature points belonging to active keyframes and visible in the new keyframe.
$Z_{cor}$ is the rescaled CNN depth.

\textbf{Sparse to Dense Filtering.} During the training we used Eq.~\ref{eq:depth_from_normal_3} to recast the depth of the points using depth and normal of other pixels around them inside a windows with predefined size.
However, this approach is error prone when the input depth image is sparse. 
For example, in a scene where a desk stands on a flat floor, the depth of the edges on the desk can be propagated from the floor since they are equally ``flat'' in the same 3D direction. If so, the depth is recomputed from a wrong parallel surface rather than the actual coplanar tangent.
To avoid this, a fast pre-segmentation into super-pixels is performed. The depth will only be filled by reprojecting from adjacent pixels within the same super-pixel. The new filtering criteria $\tilde{\vect{\mathcal{C}}}$ is defined as:
\begin{equation}
\tilde{\vect{\mathcal{C}}}_i = \{(x_j , y_j , z_j )| \vect n_j^T \vect n_i > \psi, c_i= c_j\}
\end{equation}
where $c$ is the label assigned by the super-pixel segmentation.

The overall filtering based sparse to dense reconstruction can be summarized into three steps: (1) filter $\vect Z_{opt}$ with CNN normal using Eq.~\ref{eq:depth_from_normal_3} with the new criteria $\tilde{\vect{\mathcal{C}}}$; (2) filter the updated $\vect Z_{opt}$ with bilateral filtering, the kernel weight is based on the color difference and spatial distance; (3) a wrap up filtering with CNN normal using Eq.~\ref{eq:depth_from_normal_3} with original criteria $\vect{\mathcal{C}}$. As mentioned in the previous section, the filtering is parallelized and performed on GPU, which allows us to meet the requirement of real time.

Step (1) can effectively diminish the incorrect depth reprojection. The downside of this is that it results in no value exchange between blobs. To tackle with this issue, step (2), a classical bilateral filtering is conducted. However, as the color based smoothing is not as reliable as the normal, we perform step (3) for further regularization. We denote the final reconstructed depth as $\vect{Z}_{dense}$ to distinguish it from the intermediate output $\vect{Z}_{re}$ (only used for training).

\subsection{Dense Global Fusion}

\textbf{Keyframe-wise Refinement.} In CNN-SLAM, based on LSD-SLAM \cite{LSD-SLAM}, an uncertainty based update is used for dense depth refinement. We build on DSO instead. DSO uses a window based optimization scheme, containing a bundle of active keyframes for more robust estimation. It is very expensive to associate and update the dense depth and uncertainty using every single frame. However, the depth uncertainty has already been greatly reduced by the dense reconstruction which directly propagates the low uncertainty points using the geometrical structure. In Fig.~\ref{fig:digest}, we see that a 3D reconstruction can be performed even \textbf{without} the refinement. That said, the refinement helps reject outliers and grant additional baseline stimulus in a dense manner and we therefore include it in our pipeline. However, we only perform the refinement between keyframes, and not between every frame. Specifically, we use the Bayesian Estimation based on REMODE~\cite{REMODE}\footnote{\url{https://github.com/uzh-rpg/rpg_open_remode}} and estimate the uncertainty for each pixel based on the difference between updated depth and scale fixed CNN depth (Eq.~\ref{eq:depth_optimized_2}).

As a final step in our SLAM system we deploy a fusion based method to build a global 3D model consisting of surfel splats. It is fed our refined dense depth images. Our implementation is based on  ElasticFusion~\cite{ElasticFusion}\footnote{\url{https://github.com/mp3guy/ElasticFusion}} with frame-to-frame tracking disabled since we only use it as an advanced mapper.
As the point cloud is fused into the global model, transient noise can be further rejected and loop closures are handled.

\section{Experiments}
In this section, we evaluate the effectiveness of our S2D framework on the TUM~\cite{TUM} and ICL-NUIM~\cite{ICL} RGB-D datasets. The Absolute Trajectory Error (ATE) and Percentage of Correct Depth  (PCD) (also used in~\cite{CNN-SLAM}) are used as metrics to compare with other learning/non-learning based monocular VO and SLAM systems.

The training of FCDRN is conducted using a desktop with an Intel i7-4790 processor and dual Nvidia 1080 graphic cards. The testing is done with a laptop with Intel i7-7700HQ and mobile version Nvidia 1070. 
The core of the sparse to dense reconstruction is the normal based spatial filtering together with super-pixel segmentation and color based bilateral filtering. These steps can all be greatly accelerated by GPU computing. In the experiments, we did not find a notable frame drop with our implementation in C++ with CUDA.
The running framerate of the overall system on our laptop was more than 23Hz using images of size $320\times240$. The inference time of FCDRN is around 25Hz (21Hz including copying from CPU to/back GPU) on our laptop. 

\begin{figure}[!tp]
    \centering
	\begin{subfigure}{0.4\textwidth}
		\centering
		\includegraphics[width=0.95\textwidth]{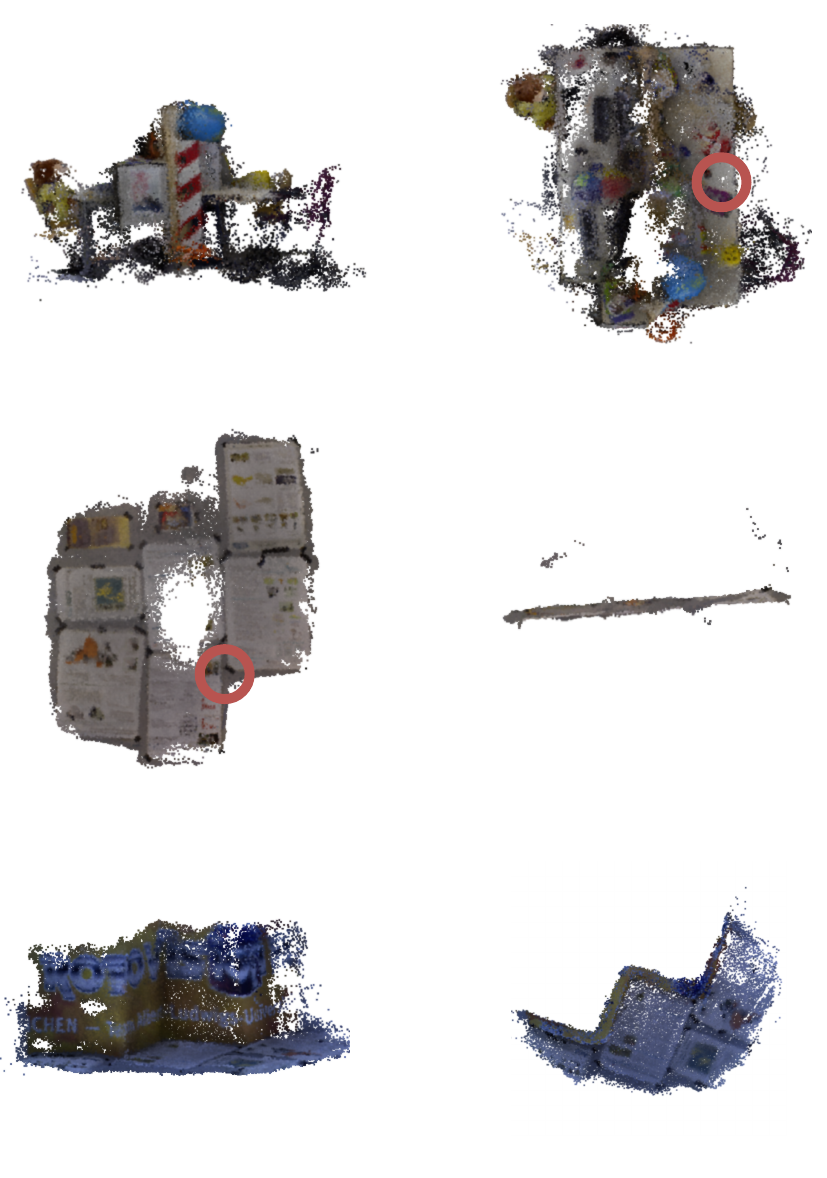}	
	\end{subfigure}
	\caption{Raw point cloud reconstruction examples of TUM seq1, 2 and 3 (top to bottom). The left and right columns show the front and top views, respectively. The red circles mark the area where loop closures are detected.}
	\label{fig:cloud_all}
\end{figure}

\subsection{Datasets}

\textbf{Training samples}  First we pre-trained FCDRN using $44624$ RGB-D frames extracted from the SUN-3D~\cite{SUN-3D} dataset. The SUN-3D dataset includes videos recorded in various typical indoor environments. 
We sampled roughly one frame per second from the video to avoid too repetitive training samples. Noisy images, e.g., mostly occluded or overexposed, cause the training to diverge. We use a standard SIFT keypoint detector to identify images that are likely to be noisy and discard images in which we find less than 50 SIFT keypoints.
Then, we trained the network with the SUN-RGBD~\cite{SUN-RGBD} dataset containing 10k refined RGB-D images which are collected from NYUv2~\cite{NYUv2}, Berkeley B3DO~\cite{B3DO} and SUN-3D~\cite{SUN-3D}. 
The training of FCDRN is implement using Pytorch. We used the Adam optimizer with learning rate $10^{-4}$ for the pre-training, and the same learning rate decayed by $2$ every $20$ epochs for the formal training. The overall training includes $20$ epochs over the pre-trained dataset we extracted and $50$ epochs using SUN-RGBD. 

\textbf{Test sequences} 
For the testing, we used the same sequences from TUM-RGBD and ICL-NUIM datasets as in the evaluation of CNN-SLAM~\cite{CNN-SLAM}. The abbreviations TUM seq1 to seq3 refer to
 {\em  long\_office\_household, nostructure\_texture\_near\_withloop} and {\em structure\_texture\_far} of sensor $fr3$, respectively. From ICL-NUIM, the first $3$ office and living room sequences are used for testing.

\subsection{Quantitative Results}

\textbf{Absolute trajectory error~(ATE).} ATE, is a well-established metric for evaluating the quality of a predicted camera trajectory. It is defined as the root mean square error between the estimated and ground-truth camera trajectories. ATE directly shows the final performance of monocular visual tracking.

Tab.~\ref{tab:ate} shows the evaluated results. We compare the performance against the original DSO and DSO using depth from FCRN, denoted as DDSO with focal length adaption as in CNN-SLAM with Eq.~\ref{eq:depth_adapt1}. These two additional baselines can help further demonstrate the influence of different depth adaptions and the quality of the scale estimation.

Firstly, it can be seen that S2D outperforms the other methods in general. Obvious lower and more stable results have been obtained with S2D in almost all the test cases. In the exception, $ICL/living1$, the depth scale estimated by FCRN is really accurate, and as a result DDSO has the lowest ATE in this case. The overall results demonstrate the high quality of the scale estimated by our network, FCDRN. Our method wrap the focal length adaption into the training and perform more effective online scale correction on the run. In contrast, DDSO using the depth and adaption with CNN-SLAM only works on par with CNN-SLAM.

\begin{table}[!tp]
	\caption{Absolute Trajectory Error}\label{tab:ate}
	\centering
	\resizebox{1.0\hsize}{!}{
		\begin{tabular}{ l | c | c | c | c | c | c | c | c }
			\hline
			\multirow{ 2}{*}{Datasets} & \multirow{ 2}{*}{S2D} & \multirow{ 2}{*}{DDSO}         & DSO & CNN-SLAM   & LSD-B & LSD & ORB & Laina \\ 
			                           &     &              & \cite{DSO}  & \cite{CNN-SLAM}   & \cite{LSD-SLAM}  & \cite{LSD-SLAM} & \cite{LSD-SLAM} & \cite{FCRN} \\ \hline \hline
			
			TUM/seq1    & \textbf{0.071}  &  0.552            &   1.221  & 0.542 & 1.717 & 1.826 & 1.206 & 0.809 \\ \hline  
			TUM/seq2    & \textbf{0.078}  &  0.203            &   0.123   & 0.243& 0.106 & 0.436 & 0.495 & 1.337 \\ \hline 
			TUM/seq3    & \textbf{0.072}  &  0.335            &   0.648  & 0.214 & 0.037 & 0.937 & 0.733 & 0.724 \\ \hline  
			ICL/office0 & \textbf{0.132}  &  0.409            &   1.118  & 0.266 & 0.587 & 0.528 & 0.430 & 0.337 \\ \hline  
			ICL/office1 & \textbf{0.131}  &  0.155            &   0.633  & 0.157 & 0.790 & 0.768 & 0.780 & 0.218 \\ \hline  
			ICL/office2 & \textbf{0.085}  &  0.456            &   0.795  & 0.213 & 0.172 & 0.794 & 0.860 & 0.509 \\ \hline 
			ICL/living0 & \textbf{0.137}  &  0.143            &   0.404  & 0.196 & 0.894 & 0.516 & 0.493 & 0.230 \\ \hline  
			ICL/living1 & 0.082           &  \textbf{0.028}   &   0.187  & 0.059 & 0.540 & 0.480 & 0.129 & 0.060 \\ \hline  
			ICL/living2 & \textbf{0.045}  &  0.162            &   0.668   & 0.323& 0.211 & 0.667 & 0.663 & 0.380 \\ \hline  
			
		\end{tabular}
	}
\end{table}

\textbf{Percentage of Correct Depth~(PCD).} PCD is defined as the percentage of depth predictions whose absolute error is smaller than $10\%$ of the ground truth depth. This reveals the quality of final depth of keyframes of our and other methods. The results are shown in Tab.~\ref{tab:dpa}. We achieve better results than CNN-SLAM in all but two of the sequences and in many the difference is large. The sequences, $ICL/office1$ and $ICL/office2$, where CNN-SLAM is better than S2D, are from the artificially refined ICL dataset. On some of the datasets, e.g., $TUM/seq1$, $TUM/seq2$ and $ICL/living2$, the PCD of our method is more than twice that of CNN-SLAM, illustrating the impact of our geometrical sparse to dense reconstruction using the geometric normal. Not surprising, S2D dramatically outperforms other classical methods shown for completeness in the table.

\subsection{Qualitative Results} 

Fig.~\ref{fig:cloud_all} shows top and front views of 3D constructions using S2D on the three TUM sequences. Shown in the top views, sharp edges are well preserved. This is accomplished by the verification via tracking of DSO and good quality of normal prediction for depth reprojecting. The original depth usually suffers from ambiguous boundaries.
Both $TUM/seq1$ and $TUM/seq2$ have loop closures that have been detected (marked with red circles in Fig.~\ref{fig:cloud_all}). These closed loops also provide evidence that the monocular depth estimation of our method is consistent. A fusion based mapper requires correct alignment between the active and the global model. On the contrary, a pose graph based key frame management approach only need a minimum of two keyframes to be aligned. 
However, this does not necessarily mean that loop closure using pose graphs is easier for monocular SLAM. Quality is what matters and the deformation model of ElasticFusion helps to achieve this goal via rejecting outliers based on the surface quality and refining appearance based on the elastic deformative graph.

\begin{table}[!tp]
	\caption{Percentage of Correct Depth}\label{tab:dpa}
	\centering
	\resizebox{1.0\hsize}{!}{
		\begin{tabular}{ l | c | c | c | c | c | c | c  }
			\hline
			\multirow{ 2}{*}{Datasets}    & \multirow{ 2}{*}{S2D} &  CNN-SLAM                    & LSD-B & LSD & ORB & Laina & REMODE \\
			 &  &  \cite{CNN-SLAM}  & \cite{LSD-SLAM} & \cite{LSD-SLAM} & \cite{ORB-SLAM2} & \cite{FCRN} & \cite{REMODE} \\ \hline \hline
			TUM/seq1    & \textbf{53.287} &  12.477          & 3.797 & 0.086 & 0.031 & 12.982 &  9.548 \\ \hline 
			TUM/seq2    & \textbf{66.628} &  24.077          & 3.966 & 0.882 & 0.059 & 15.412 & 12.651 \\ \hline
			TUM/seq3    & \textbf{37.683} &  27.396          & 6.449 & 0.035 & 0.027 & 9.450  &  6.739 \\ \hline 
			ICL/office0 & \textbf{27.445} &  19.410          & 0.603 & 0.335 & 0.018 & 17.194 &  4.479 \\ \hline 
			ICL/office1 & 19.702          &  \textbf{29.150} & 4.759 & 0.038 & 0.023 & 20.838 &  3.132 \\ \hline 
			ICL/office2 & 27.059          &  \textbf{37.226} & 1.435 & 0.078 & 0.040 & 30.639 & 16.708 \\ \hline
			ICL/living0 & \textbf{19.337} &  12.840 & 1.443 & 0.360 & 0.027 & 15.008 &  4.479 \\ \hline 
			ICL/living1 & \textbf{25.090} &  13.038 & 3.030 & 0.057 & 0.021 & 11.449 &  2.427 \\ \hline 
			ICL/living2 & \textbf{68.907} &  26.560 & 1.807 & 0.167 & 0.014 & 33.010 &  8.681 \\ \hline 
			
		\end{tabular}
	}
\end{table}

\subsection{Discussion}
In this subsection, we discuss the limitations of S2D. 
The overall performance of S2D relies on two major factors: the performance of visual tracking and the generalization ability of FCDRN for predicting depth/normals. 

In the supplementary material, we present the ATE of S2D for all available TUM/ICL sequences to allow for future comparisons to S2D. We compare it with the results from ElasticFusion~\cite{ElasticFusion} which uses the captured depth image as input. 
S2D works well in general, but for some sequences, both S2D and ElasticFusion lose track. The reasons reported in \cite{ElasticFusion} are valid also for S2D. The sequences in question exhibit a high rate of dropped frames and sudden high angular velocities, which mainly affect the image alignment and thus the tracking. 
As for the FCDRN, it struggles to generate accurate depth priors when the input image is close to textureless. These two challenges can be addressed, e.g., by using a global shutter camera with higher frame rate~\cite{DSO} and by combining the visual input with inertial data~\cite{VI-DSO}. 

S2D has been developed for indoor use. To investigate how S2D generalizes to outdoor scenes we performed tests with the KITTI odometry dataset~\cite{kitti}. 
Tab.~\ref{tab:KITTI} shows the translational, $t_{rel}(ratio)$, and rotational, $r_{rel}(10^{-3} \times deg/m)$, RMSEs for some sequences. The official tool provided by \cite{kitti} is used for the evaluation. 
We compared three methods: original DSO, S2D with FCDRN trained by the SUN dataset (indoor as above) and S2D with FCDRN fine-tuned using the same KITTI raw sequences as in \cite{Left-right}. The CNN depth is fine-tuned with the same loss function as in \cite{semo,DVSO}. The CNN normal is fine-tuned based on the coupled refinement term from Eq.~\ref{eq:loss_all} as ground truth depth is not available to calculate the normals. Note that we only finetune the CNN in S2D and do not change anything else in the system pipeline.
Thanks to the robust tracking of DSO, the rotational errors are low for all three methods (note the scale of $r_{rel}$). On the other hand, there are significant differences in translational errors. S2D trained on indoor scenes is not working so well outdoors, as can be expected. We still see a clear improvement over DSO. 
When the CNN in S2D is fine-tuned for the outdoor environment, the translational errors are significantly reduced. The $t_{rel}$ is now on par with results of DSO with state-of-the-art depth priors for single view images, achieving an average of $0.107$ according to~\cite{DVSO}. 
To fully convert S2D from indoor use to outdoor use, one should take into account that stereo data is the main source for training in outdoor environments, in contrast to indoor scenes where RGB-D data dominates. One could redesign FCDRN to additionally predict a virtual stereo pair and incorporate the impressive results of DVSO~\cite{DVSO}. 
However, we want to emphasize that the focus in this paper is utilizing the learning based geometrical information to densely reconstruct scenes from corrected sparse depth and that we target indoor scenes.

\begin{table}[!tp]
\caption{RMSEs on KITTI odometry dataset}\label{tab:KITTI}
\centering

	\resizebox{0.8\hsize}{!}{
\begin{tabular}{ c | c | c | c  | c | c| c}
	\hline
	& 	\multicolumn{2}{c|}{DSO} & 	\multicolumn{2}{c|}{S2D}  & \multicolumn{2}{c}{S2D fine-tuned}  \\ \hline
	Sequence No.	& $t_{rel}$ &	$r_{rel}$ &	$t_{rel}$	& $r_{rel}$	& $t_{rel}$ & $r_{rel}$ \\ \hline \hline      
	00	& 0.487 &	0.046 &	0.213	& 0.058	& 0.107 & 0.055 \\ \hline
	02	& 0.640 &	0.040 &	0.211	& 0.044	& 0.089 & 0.045 \\ \hline
	04	& 0.979 &   0.035 & 0.706	& 0.021	& 0.035 & 0.015  \\ \hline
	06	& 0.571 &	0.111 &	0.136	& 0.186	& 0.096 & 0.129  \\ \hline
	08	& 0.570 &	0.084 &	0.234	& 0.099	& 0.077 & 0.092  \\ \hline          
	mean & 0.649 &	0.063 &	0.300	& 0.082	& 0.081 & 0.067  \\ \hline  
\end{tabular}
}
\end{table}

\section{Summary and Conclusions}

In this paper, a new deep learning based monocular SLAM method is proposed. A single CNN has been trained to predict depth and normals in a coupled way. The depth is used in the projective geometric optimization for accurate pose estimation. The normals are utilized for a dense geometrical reconstruction using intermediate sparse optimized point clouds. Experiments demonstrated the effectiveness of our method, S2D. Both improved motion estimation and dense depth reconstruction are achieved in comparison with state-of-the-art deep dense monocular SLAM.

In future work, we plan to investigate including the camera pose estimation in the depth/normal training scheme.
S2D is not limited to mapping with a single camera, it can potentially be used for reconstruction with multiple sensors having sparse depth measurements, e.g., camera and Lidar. We also plan to investigate how to support human interaction with the dense 3D reconstruction. To achieve this goal, we will study exploiting the semantics of the environment in the model.

{\small
	\balance
	\bibliographystyle{IEEEtran}
	\bibliography{bib}
}

\end{document}


\maketitle
	\thispagestyle{empty}
	\pagestyle{empty}
	
\section{Introduction}
In this supplementary material, we include additional experimental results for our paper.
Firstly, we list ATE tests results on all TUM and ICL datasets. Then, we show different intermediate outputs from S2D. 

\section{Additional results on ICL and TUM datasets}
	Tab.~\ref{tab:ICL} shows the ATE results for all ICL sequences and Tab.~\ref{tab:TUM} for all TUM sequences. Results from ElasticFusion[39] using ground truth depth as input are listed as reference. 
	Note that CNN-SLAM was only evaluated on the sequences that we include in the paper.
	
\begin{table}[!hp]
	\centering
	
	\begin{tabular}{ c | c | c   }
		\hline
		Sequence Name  & 	S2D	    & Elastic Fusion  \\ \hline \hline           
		living room0	 & 0.137	& 0.009 \\ \hline           
		living room1	 & 0.082	& 0.009 \\ \hline           
		living room2	 & 0.045	& 0.014 \\ \hline           
		living room3	 & 0.067	& 0.106 \\ \hline           
		office room0	 & 0.132	& NaN \\ \hline           
		office room1	 & 0.131	& NaN \\ \hline           
		office room2	 & 0.085	& NaN \\ \hline           
		office room3	 & 0.094	& NaN \\ \hline           
		
	\end{tabular}
	\caption{Absolute Trajectory Error on ICL dataset.}\label{tab:ICL}
\end{table}

\begin{table}[!hp]
\centering
\resizebox{0.5\hsize}{!}{
\begin{tabular}{ c | c | c | c  }
	\hline
	Sensor    &  Sequence Name                       &   S2D       & Elastic Fusion \\ \hline  \hline          
	freiburg1  &  360                                 &   0.161     & 0.108 \\ \hline  
	freiburg1  &  desk                                &   0.126     & 0.02 \\ \hline   
	freiburg1  &  desk2                               &   0.090     & 0.048 \\ \hline   
	freiburg1  &  floor                               &   --        & -- \\ \hline 
	freiburg1  &  plant                               &   0.141     & 0.022 \\ \hline    
	freiburg1  &  room                                &   0.221     & 0.068 \\ \hline   
	freiburg1  &  rpy                                 &   0.043     & 0.025 \\ \hline  
	freiburg1  &  teddy                               &   0.401     & 0.083 \\ \hline    
	freiburg1  &  xyz                                 &   0.030     & 0.011 \\ \hline 
	freiburg2  &  360 hemisphere                      &   1.053     & -- \\ \hline             
	freiburg2  &  360 kidnap                          &   --         & -- \\ \hline      
	freiburg2  &  coke                                &   --         & -- \\ \hline  
	freiburg2  &  desk                                &   0.162     & 0.071 \\ \hline   
	freiburg2  &  dishes                              &   0.190      & -- \\ \hline    
	freiburg2  &  large no loop                       &   0.914     & -- \\ \hline            
	freiburg2  &  large with loop                     &   --         & -- \\ \hline          
	freiburg2  &  metallic sphere                     &   0.422     & -- \\ \hline              
	freiburg2  &  metallic sphere2                    &   0.314     & -- \\ \hline               
	freiburg2  &  pioneer 360                         &   --         & -- \\ \hline      
	freiburg2  &  pioneer slam                        &   --         & -- \\ \hline       
	freiburg2  &  pioneer slam2                       &   --         & -- \\ \hline        
	freiburg2  &  pioneer slam3                       &   --         & -- \\ \hline        
	freiburg2  &  rpy                                 &   0.023     & 0.015 \\ \hline  
	freiburg2  &  xyz                                 &   0.008     & 0.011 \\ \hline  
	freiburg3  &  cabinet                             &   0.527     & -- \\ \hline      
	freiburg3  &  large cabinet                       &   0.107     & 0.099 \\ \hline            
	freiburg3  &  long office household               &   0.071     & 0.017 \\ \hline                    
	freiburg3  &  nostructure notexture far           &   --         & -- \\ \hline                    
	freiburg3  &  nostructure notexture near withloop &   --         & -- \\ \hline                              
	freiburg3  &  nostructure texture far             &   0.693     & 0.074 \\ \hline                      
	freiburg3  &  nostructure texture near withloop   &   0.078     & 0.016 \\ \hline                                
	freiburg3  &  structure notexture far             &   0.179     & 0.03 \\ \hline                      
	freiburg3  &  structure notexture near            &   0.853     & 0.021 \\ \hline                       
	freiburg3  &  structure texture far               &   0.072     & 0.013 \\ \hline                    
	freiburg3  &  structure texture near              &   0.134     & 0.015 \\ \hline                     
	freiburg3  &  teddy                               &   0.039     & 0.049 \\ \hline    
	\end{tabular}
	}
\caption{Absolute Trajectory Error on the full TUM dataset.}
\label{tab:TUM}
\end{table}

\clearpage
\section{Intermediate Results}
    
\begin{figure}[!hp]
	\centering
	\begin{subfigure}{0.6\textwidth}
		\centering
		\includegraphics[trim={8cm 4cm 8cm 4cm},clip,width=\textwidth]{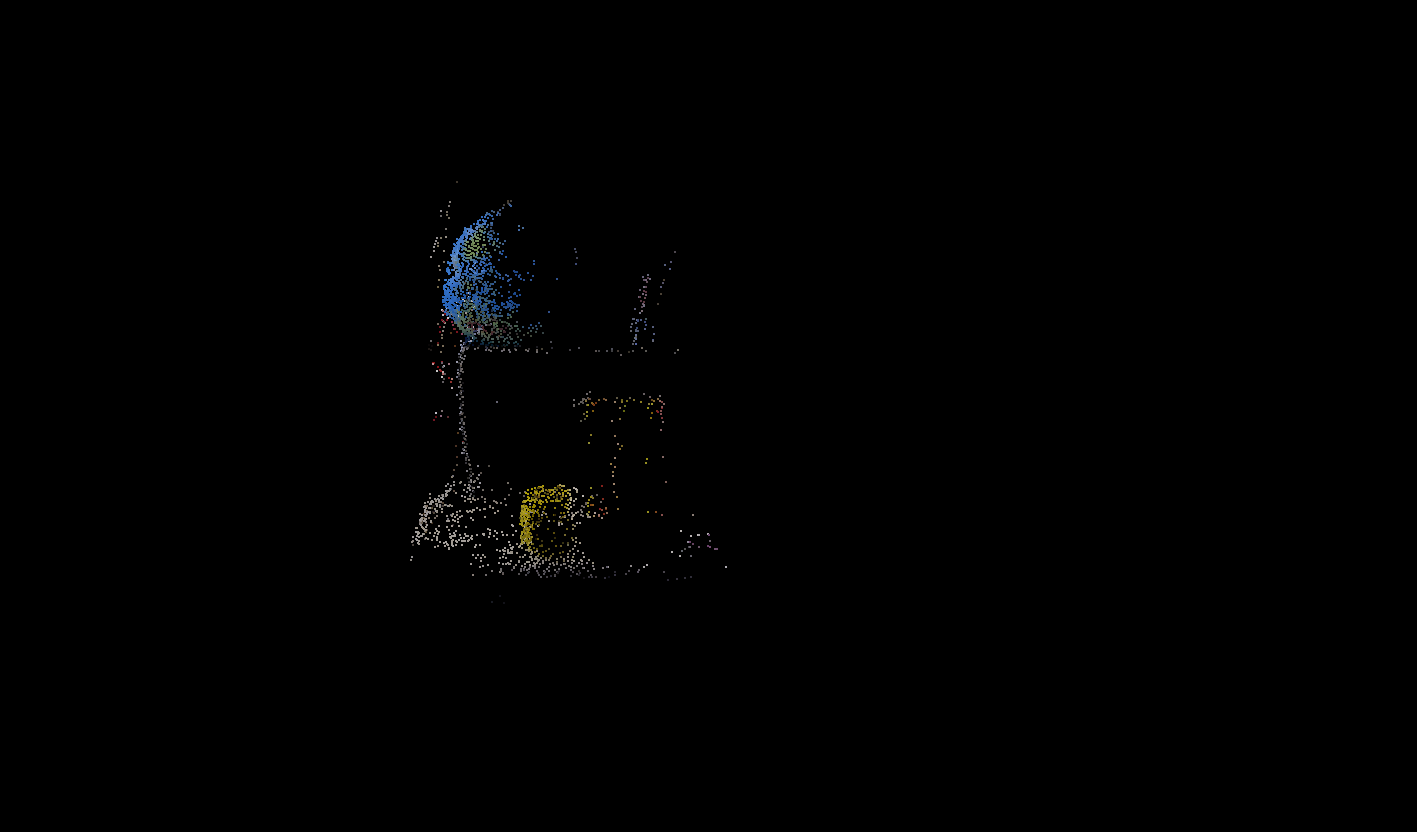}
		\caption{Rescaled CNN depth,  $\vect{Z}_{cor}$.}
	\end{subfigure}
	\hfill
	\begin{subfigure}{0.6\textwidth}
		\centering
		\includegraphics[trim={8cm 4cm 8cm 4cm},clip,width=\textwidth]{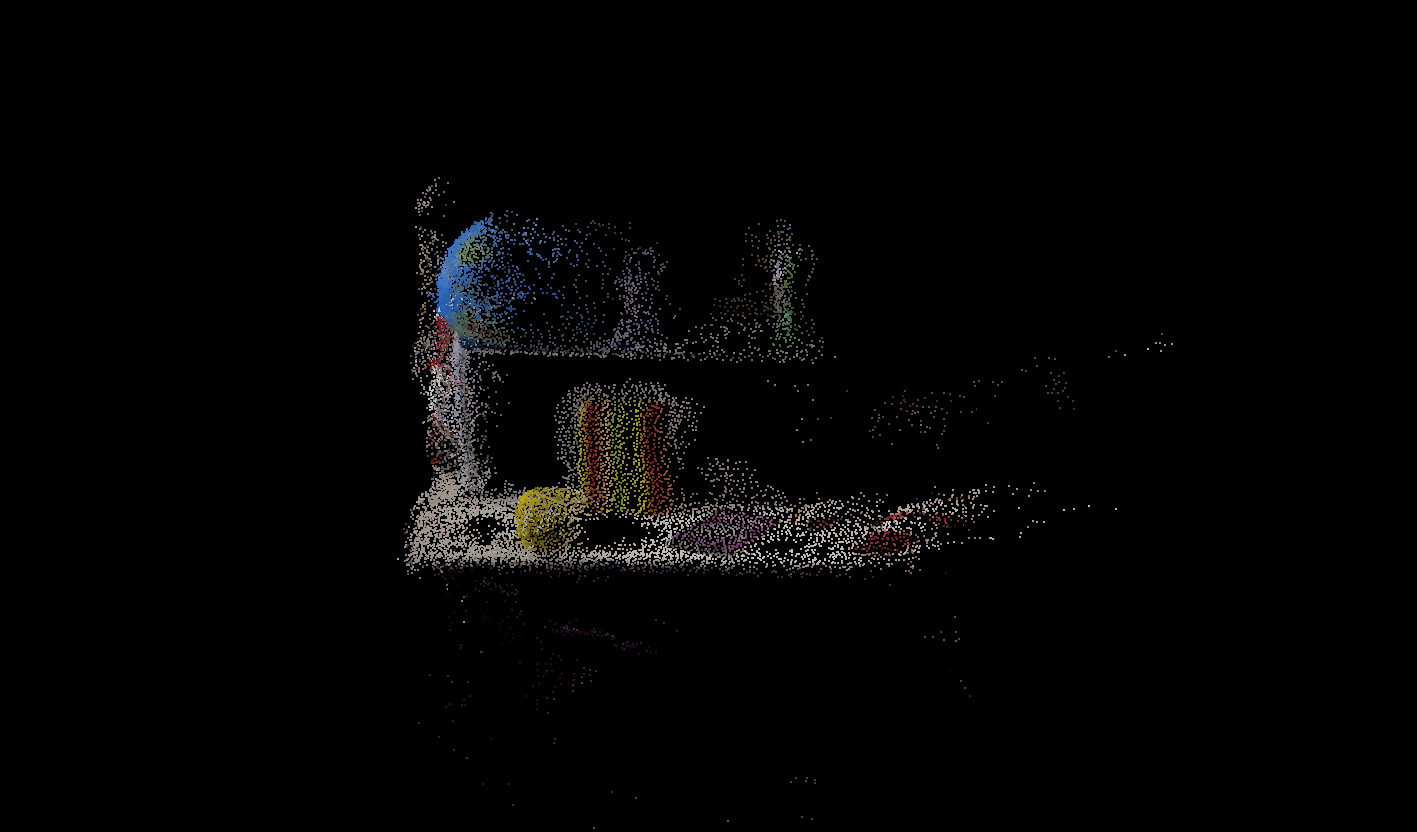}
		\caption{Reconstructed dense depth,  $\vect{Z}_{dense}$.}
	\end{subfigure}
	\caption{Comparison of 3D reconstructions using different source of depth as input to the mapping backend. Top:  $\vect{Z}_{cor}$ complemented with further updated depths from $\vect{Z}_{opt}$. Bottom: Our final reconstructed depth, $\vect{Z}_{dense}$.}
	\label{fig:cloud_compare}
\end{figure}

	Fig.~\ref{fig:cloud_compare} shows a comparison of 3D constructions using rescaled CNN depth, $\vect{Z}_{cor}$, and reconstructed depth, $\vect{Z}_{dense}$. In addition, the sparse optimized points $\vect{Z}_{opt}$ are already updated in $\vect{Z}_{cor}$. Both depth resulting images have been through the refinement. It shows us that the CNN depth, $\vect{Z}_{cor}$, suffers from inconsistencies between frames. The resulting merged 3D model is sparse since a lot of points have been rejected as transient noises. In contrast, sparse to densely reconstructed depth, $\vect{Z}_{dense}$, merges easily resulting in a higher density model. This also shows that S2D is not mainly relying on the refinement since the Bayesian estimation refinement is used in both cases.
	
\begin{figure*}[!hp]
	\centering
	\begin{subfigure}{0.45\textwidth}
		\centering
		\includegraphics[width=\textwidth]{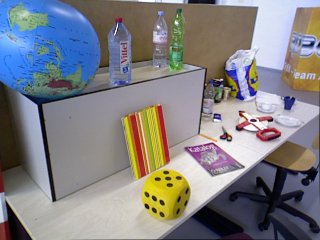}
		\caption{Color image.}
	\end{subfigure}
	\hfill
	\centering
	\begin{subfigure}{0.45\textwidth}
		\centering
		\includegraphics[width=\textwidth]{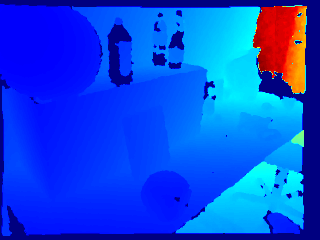}
		\caption{Ground truth depth image.}
	\end{subfigure} 
	\centering
	\begin{subfigure}{0.45\textwidth}
		\centering
		\includegraphics[width=\textwidth]{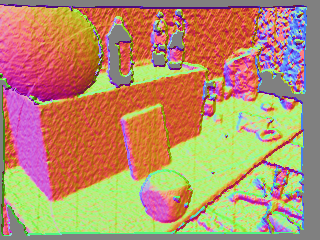}
		\caption{Normal from Ground truth depth.}
	\end{subfigure}
	\hfill
	\centering
	\begin{subfigure}{0.45\textwidth}
		\centering
		\includegraphics[width=\textwidth]{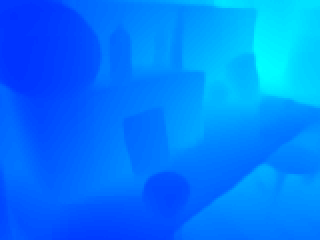}
		\caption{CNN depth.}
	\end{subfigure}
	\centering
	\begin{subfigure}{0.45\textwidth}
		\centering
		\includegraphics[width=\textwidth]{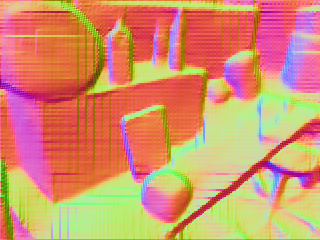}
		\caption{Normal from CNN depth.}
	\end{subfigure}
	\hfill
	\centering
	\begin{subfigure}{0.45\textwidth}
		\centering
		\includegraphics[width=\textwidth]{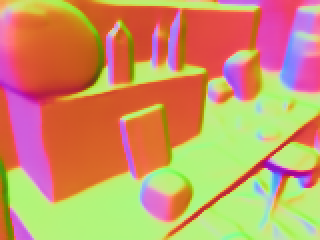}
		\caption{CNN normal.}
	\end{subfigure}
	\caption{Depth/normal comparison I.}
	\label{fig:set1}
\end{figure*}

\begin{figure*}[!hp]
	\centering
	\begin{subfigure}{0.45\textwidth}
		\centering
		\includegraphics[width=\textwidth]{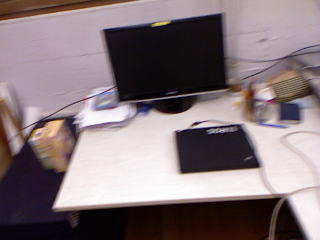}
		\caption{Color image.}
	\end{subfigure}
	\hfill
	\centering
	\begin{subfigure}{0.45\textwidth}
		\centering
		\includegraphics[width=\textwidth]{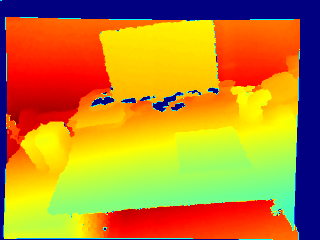}
		\caption{Ground truth depth image.}
	\end{subfigure}
	\centering
	\begin{subfigure}{0.45\textwidth}
		\centering
		\includegraphics[width=\textwidth]{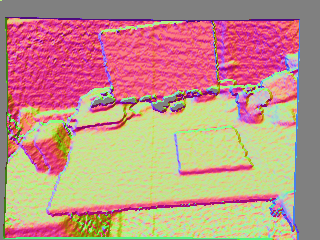}
		\caption{Normal from ground truth depth.}
	\end{subfigure}
	\hfill
	\centering
	\begin{subfigure}{0.45\textwidth}
		\centering
		\includegraphics[width=\textwidth]{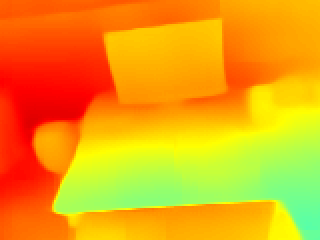}
		\caption{CNN depth.}
	\end{subfigure}
	\centering
	\begin{subfigure}{0.45\textwidth}
		\centering
		\includegraphics[width=\textwidth]{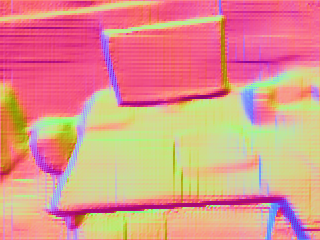}
		\caption{Normal from CNN depth.}
	\end{subfigure}
	\hfill
	\centering
	\begin{subfigure}{0.45\textwidth}
		\centering
		\includegraphics[width=\textwidth]{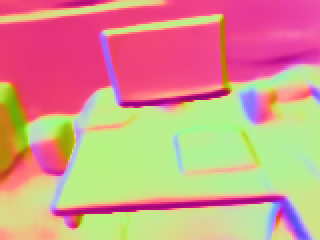}
		\caption{CNN normal.}
	\end{subfigure}
	\caption{Depth/normal comparison II.}
	\label{fig:set2}
\end{figure*}

	Fig.~\ref{fig:set1} and Fig.~\ref{fig:set2} show intermediate results from S2D. More precisely, they show a captured RGB-D image, CNN predicted depth/normal, normal images computed from CNN depth and captured depth using Eq.3. Thanks to the coupled training, the CNN depth are naturally smooth without employing any explicit smoothing terms in the loss like other existing methods. The CNN normal are clearly of better quality compared with the one computed from CNN depth.
	
